\documentclass[letterpaper]{article} % DO NOT CHANGE THIS
\usepackage{aaai25}  % DO NOT CHANGE THIS
\usepackage{times}  % DO NOT CHANGE THIS
\usepackage{helvet}  % DO NOT CHANGE THISp
\usepackage{courier}  % DO NOT CHANGE THIS
\usepackage[hyphens]{url}  % DO NOT CHANGE THIS
\usepackage{graphicx} % DO NOT CHANGE THIS
\usepackage{booktabs}

\usepackage{natbib}  % DO NOT CHANGE THIS AND DO NOT ADD ANY OPTIONS TO IT
\usepackage{caption} % DO NOT CHANGE THIS AND DO NOT ADD ANY OPTIONS TO IT
\usepackage{amsthm,amsmath,amssymb}
\usepackage{newfloat}
\usepackage{listings}
\usepackage{subfigure}
\usepackage{enumitem}
\usepackage{pifont}
\usepackage{float}
\usepackage{multirow}
\usepackage[ruled]{algorithm2e}
\usepackage{color}
\usepackage{makecell}
\newcommand{\modelname}{CurvGIB}
\newtheorem{myDef}{Definition}

\newtheorem{myPro}{Proposition}

\usepackage[toc,title,page]{appendix}
\usepackage[table,xcdraw]{xcolor}
\usepackage{newfloat}
\usepackage{listings}
\DeclareCaptionStyle{ruled}{labelfont=normalfont,labelsep=colon,strut=off} % DO NOT CHANGE THIS
\floatstyle{ruled}
\newfloat{listing}{tb}{lst}{}
\floatname{listing}{Listing}
\title{Discrete Curvature Graph Information Bottleneck}
\author{
    Xingcheng Fu\textsuperscript{\rm 1, \rm 2}~\thanks{Co-corresponding Author.}, 
    Jian Wang\textsuperscript{\rm 1, \rm 2}, 
    Yisen Gao\textsuperscript{\rm 3}, 
    Qingyun Sun\textsuperscript{\rm 4 *}, 
    Haonan Yuan\textsuperscript{\rm 4}, \\
    Jianxin Li\textsuperscript{\rm 4}, 
    Xianxian Li\textsuperscript{\rm 1, \rm 2}
}
\affiliations{
    \textsuperscript{\rm 1}Guangxi Key Lab of Multi-source Information Mining \& Security, Guangxi Normal University, Guilin, China\\
    \textsuperscript{\rm 2}Key Lab of Education Blockchain and Intelligent Technology, Ministry of Education, Guangxi Normal University, China\\
    \textsuperscript{\rm 3}Institute of Artificial Intelligence, Beihang University, Beijing, China\\
    \textsuperscript{\rm 4}School of Computer Science and Engineering, Beihang University, Beijing, China\\
    \{fuxc, lixx\}@gxnu.edu.cn, wj668811@stu.gxnu.edu.cn, \{yisengao, sunqy, yuanhn, lijx\}@buaa.edu.cn
}

\usepackage{bibentry}
\begin{document}

\maketitle

\begin{abstract}
Graph neural networks(GNNs) have been demonstrated to depend on whether the node effective information is sufficiently passing. 
Discrete curvature (\textit{Ricci curvature}) is used to study graph connectivity and information propagation efficiency with a geometric perspective, and has been raised in recent years to explore the efficient message-passing structure of GNNs. 
However, most empirical studies are based on directly observed graph structures or heuristic topological assumptions and lack in-depth exploration of underlying optimal information transport structures for downstream tasks. 
We suggest that graph curvature optimization is more in-depth and essential than directly rewiring or learning for graph structure with richer message-passing characterization and better information transport interpretability. 
From both graph geometry and information theory perspectives, we propose the novel Discrete \textbf{Curv}ature \textbf{G}raph \textbf{I}nformation \textbf{B}ottleneck (\textbf{\modelname}) framework to optimize the information transport structure and learn better node representations simultaneously. 
\modelname~advances the \textit{Variational Information Bottleneck} (\textit{VIB}) principle for Ricci curvature optimization to learn the optimal information transport pattern for specific downstream tasks. 
The learned \textit{Ricci curvature} is used to refine the optimal transport structure of the graph, and the node representation is fully and efficiently learned. 
Moreover, for the computational complexity of Ricci curvature differentiation, we combine \textit{Ricci flow} and \textit{VIB} to deduce a curvature optimization approximation to form a tractable IB objective function. 
Extensive experiments on various datasets demonstrate the superior effectiveness and interpretability of \modelname.  
\end{abstract}
\section{Introduction}
% Graph Neural Networks (GNNs) have made significant progress in graph learning in recent years and have shown broad application prospects in various fields such as social network analysis, recommendation systems, and bio-informatics. 
Through the unique mechanism of message passing, graph neural networks (GNNs) are capable of effectively capturing complex relationships and structural information between nodes.
Extensive empirical studies have proved that whether the node information can be sufficient and effectively propagated is crucial, and the efficiency of message passing directly impacts the learning capability of GNNs. 
For different downstream tasks and applications, studying the optimal message passing structure in GNNs is of crucial necessity and importance for improving their overall performance and adaptability.

\begin{figure}[t] % 'htbp' options for positioning
    \centering
    \includegraphics[width=0.4\textwidth]{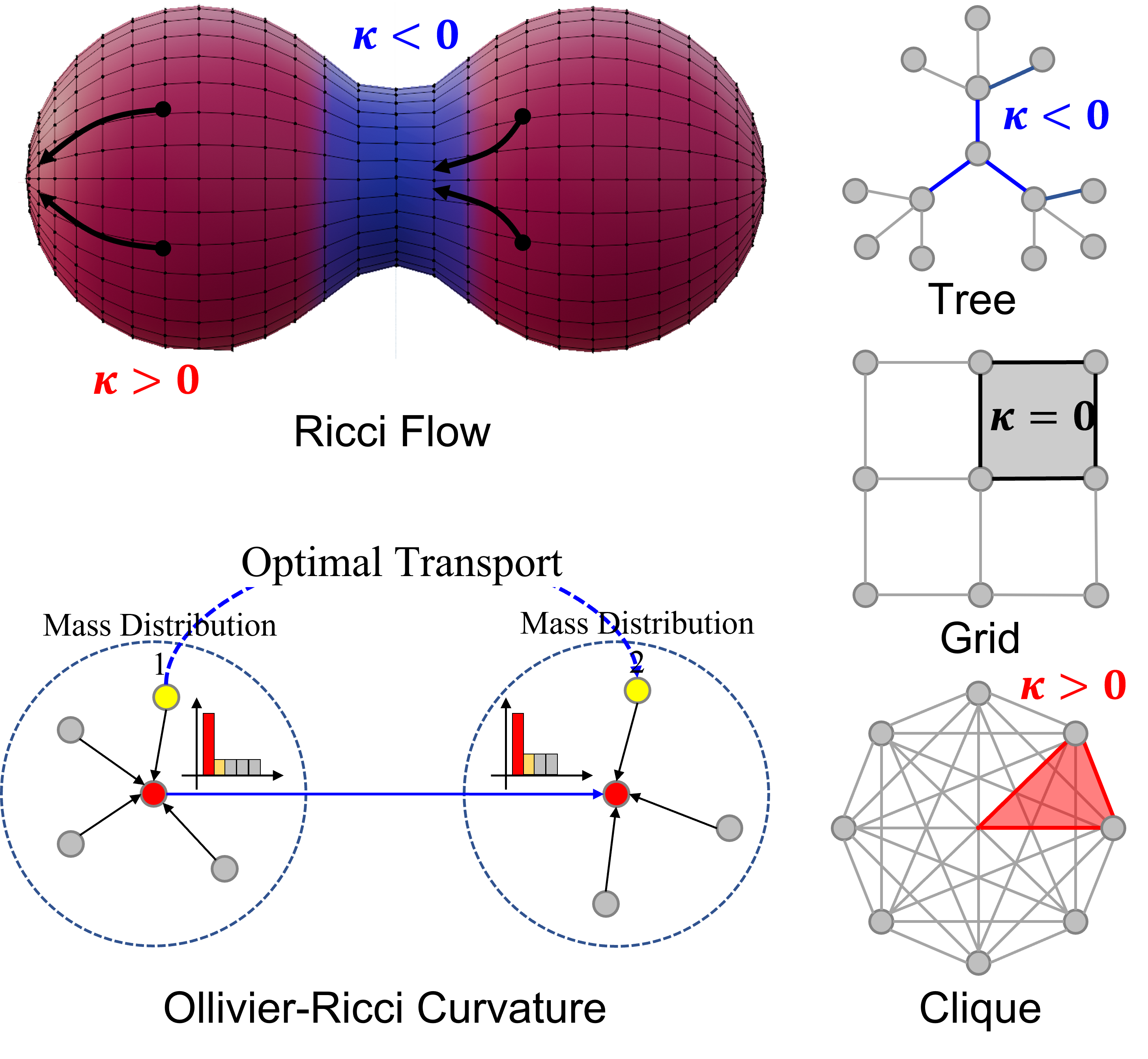}
    \caption{Ricci flow, Ollivier-Ricci curvature and different curvature-relevant topologies. }
    \vspace{-1em}
    \label{fig:intro}
\end{figure}

In practice, the properties of graphs are often determined by critical structures relevant to downstream tasks, rather than the entire graph structure~\cite{Sun2021GraphSL}. 
In response to this observation, two main research directions have emerged. 
Some of the existing work focuses on the exploration and mining of critical subgraph structures~\cite{Yu2020GraphIB} relevant to tasks by leveraging some information-theoretic techniques, such as maximum mutual information~\cite{Wu2020GraphIB} and information bottleneck~\cite{tishby2000information}. 
On the other hand, other works focus on the messaging perspective, and these works are devoted to addressing messaging issues such as excessive smoothing and excessive compression, which can seriously impair GNNs' performance.
Recently, the definition of graph curvature has provided a new perspective and insight for understanding and alleviating technical bottlenecks in the message-passing mechanism~\cite{topping2021understanding,Nguyen2022RevisitingOA,sun2024riccinet}. 
However, the first existing works only focuses on finding task-relevant information and lacks a deep understanding of the message-passing process in GNNs; the second existing works relies primarily on heuristic graph geometric priors based on the original topology to guide structure re-wiring, making it difficult to bridge and correlate with downstream tasks. 
There is still a lack of a unified perspective to comprehensively understand and mine the underlying task-relevant optimal transport structures in GNNs. 

Rethinking the above issues, we suggest that the representation and generalization of GNNs fundamentally depend on whether the underlying transport structure of effective information is optimal in the broad and different downstream task scenarios.  
To extract task-relevant information more efficiently, the Information Bottleneck (IB) principle provides us with a theoretical framework, guiding us on how to maximize the retention of task-relevant information while compressing irrelevant information. 
As shown in Figure~\ref{fig:intro}, graph curvature as a geometric definition of optimal transport, offers intuitive and effective metrics for the connectivity patterns of local topological structures. 
Consequently, a natural question arises: \textit{\textbf{Can we combine the perspectives of information theory and graph geometry to develop a novel framework that can learn the optimal transport of underlying task-relevant effective information? }}
This unified framework is expected to provide important guidance for our deep understanding of the underlying transmission structure of effective information in GNNs, and further broadening the application of GNNs in various practical scenarios. 

In this paper, we present a unified discrete \textbf{\underline{Curv}}ature \textbf{\underline{G}}raph \textbf{\underline{I}}nformation \textbf{\underline{B}}ottleneck (\textbf{\modelname}) framework through both information theory and graph geometry perspectives to mining the underlying optimal transport structure of task-relevant graph information. 
The core idea is to leverage IB to learn the downstream task-relevant graph curvature rather than the topological structure directly, as the graph curvature better reflects the local connection patterns of the graph. 
Furthermore, we utilize graph curvature to distill the underlying optimal transport structure of task-relevant information, serving as a new structure for Graph Neural Networks (GNNs) training. 
Additionally, to address the issue of high computational complexity in discrete graph curvature differentiation, we improve an approximate optimization algorithm for Ricci curvature within our framework based on Ricci flow and employ a bi-level optimization technique to ensure efficient and stable learning of IB and Ricci curvature simultaneously. 
Our contributions are summarized as follows: 
\begin{itemize}[leftmargin=*]
    \item 
    \modelname~advances both the Information Bottleneck principle and discrete graph curvature for graph learning, providing an elegant and universal framework in the perspectives of information theory and graph geometry.  
    \item \modelname~provide strong geometric intuition and interpretability, and it can mine the underlying optimal transport structure of GNNs to improve the model performance simultaneously, by using the IB principle to learn the optimal graph curvatures.
    \item Extensive experiment results in node classification demonstrate that the proposed \modelname~has superior effectiveness compared to other baselines. 
\end{itemize}

\section{Related Work}

% \subsection{Graph Neural Networks}
% Building effective Graph Neural Networks (GNNs) is crucial for learning from graph data. Initially, Graph Convolutional Networks (GCNs) and Graph Attention Networks (GATs) improved the ability to capture complex relationships by combining node features with graph structure, but they face the issue of oversmoothing. Later, GraphSAGE and Graph Isomorphism Networks (GINs) addressed scalability and feature expressiveness with techniques like subgraph sampling and feature mapping. However, these methods often remain limited to local neighbors and struggle to capture global information.
\begin{figure*}[t] % 'htbp' options for positioning
    \centering
    \includegraphics[width=0.95\textwidth]{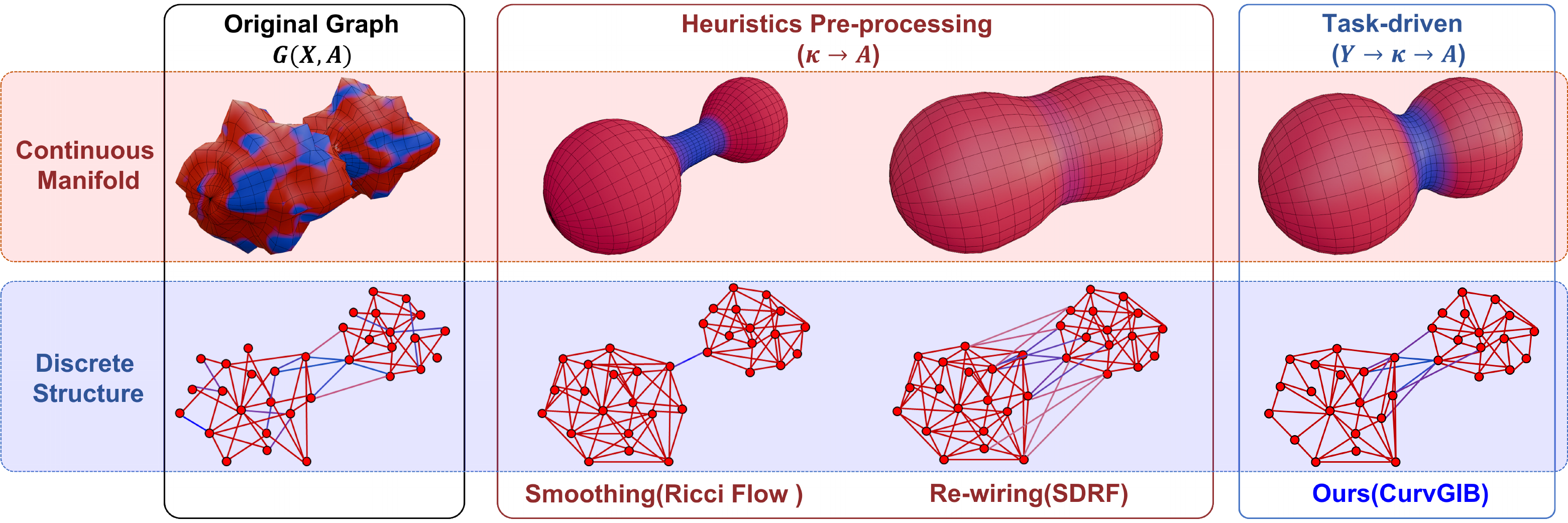}
    \vspace{-1em}
    \caption{Instances of the geometric perspective, comparison different paradigms for graph structure refinement. }
    \label{fig:flows}
    \vspace{-1em}
\end{figure*}
\subsection{Graph Structure Learning}
Graph Structure Learning (GSL) parameterizes the adjacency matrix using probabilistic models, fully parameterized models, or metric learning models and jointly optimizes the adjacency matrix and GNN parameters for downstream tasks. For example, Yixin Liu et al.\cite{Liu2022TowardsUD} use contrasting learning to generate optimal graph structures, maximizing the consistency between the learned topology and self-enhancing learning objectives. Zhen Zhang et al.\cite{Zhang2021HierarchicalMG} propose a multi-view pooling operator, which enhances the robustness of graph structures and preserves low-level topological information. However, GSL methods often face high time complexity, limiting their scalability.

\subsection{Graph Curvature}
Discrete curvature methods can be categorized into two approaches: \textbf{re-weighting} and \textbf{re-wiring}. The \textbf{re-weighting} approach influences the information propagation path by adjusting the weights of nodes or edges. 
BORF\cite{Nguyen2022RevisitingOA} uses the Ollivier-Ricci curvature to reconstruct the neighborhoods of the paired nodes, but its approximation can amplify the impact of noisy edges. 
% BORF\cite{} uses the Ollivier-Ricci curvature to reconstruct the neighborhoods of the paired nodes, but its approximation can amplify the impact of noisy edges. 
SDRF\cite{topping2021understanding} employs balanced Forman curvature for neighborhood reconstruction, but it has weaker handling of nodes in graph boundary regions and involves higher computational costs. The \textbf{re-wiring} approach optimizes information transmission by directly modifying the graph structure. Some works \cite{Ye2020CurvatureGN,fu2021ace, fu2023hyperbolic} has innovatively incorporated graph curvature information into GNNs, significantly enhancing the performance of graphs with complex structures. Despite the potential shown by these methods, most existing research focuses on static graphs, with limited exploration of dynamic or large-scale graphs.

\subsection{Information Bottleneck for Graph Learning}
As a graph compression theory, information bottleneck has been widely concerned in recent years, and a lot of work attempts to use information bottlenecks to obtain the most sufficient representation for various graph-related downstream tasks.
% In terms of optimizing graph structure, and a lot of work attempts to use information bottlenecks to obtain the most sufficient representation to improve the model's performance for various graph-related downstream tasks.
For example,\cite{Wu2020GraphIB} proposes GIB which achieves the most efficient node representation by maximizing mutual information between representations and labels while minimizing mutual information between representations and original inputs.
With a variational IB principle, it can obtain sufficient information for node representation learning.
\cite{Yu2020GraphIB} proposes SIB which directly reveals the vital substructure in the subgraph level. 
It leverages the mutual information estimator from the Deep Variational Information Bottleneck (\textit{VIB}) \cite{Alemi2017DeepVI} for irregular graph data as the GIB objective.
It tries to compress the original structure and help nodes aggregate the most efficient information coming from its neighborhood.
\cite{Sun2021GraphSL} proposes VIB-GSL which can extract the underlying relations from the essence of representation learning, which is related to downstream tasks. 
It generates new structures by sampling original features with a specific probability distribution to mask original structures.

\section{Prelimilary}
\subsection{Information Bottleneck}
Following standard practice in the IB literature~\cite{tishby2000information}, given data $X$, representation $Z$ of $X$ and target $Y$, $(X,Y,Z)$ are following the Markov Chain $<Y\to X \to Z>$. 
\begin{myDef}[Information Bottleneck]
For the input data $X$ and its label $Y$,  the \textbf{Information Bottleneck} principle aims to learn the minimal sufficient representation $Z$:
\begin{equation}\label{Eq:IB}
    Z = \arg\min_{Z}-I(Z, Y)+\beta I(Z, X),
\end{equation}
where $\beta$ is the Lagrangian multiplier trading off sufficiency and minimality. 
\end{myDef}

\textit{\textbf{Variational Information Bottleneck (VIB)}}~\cite{Alemi2017DeepVI} proposed a variational approximation of the IB objective for deep neural networks:
\begin{equation}
\begin{aligned}
    \mathcal{L}=\frac{1}{N}\sum^{N}_{i=1}\int &dZp(Z|X_i)\log q(Y_i|Z)\\
    &+\beta \mathcal{D}_{KL}\left(p(Z|X_i),r(Z)\right),
\end{aligned}
\end{equation}
% \begin{equation}
%     \mathcal{L}=\frac{1}{N}\sum^{N}_{i=1}\int dZp(Z|X_i)\log q(Y_i|Z)+\beta \mathcal{D}_{KL}(p(Z|X_i),r(Z)),   
% \end{equation}
where $q(Y_i|Z)$ is the variational approximation to $p(Y_i|Z)$ and $r(Z)$ is the variational approximation of $Z$'s prior distribution. 
\subsection{Discrete Graph Curvature}
The overlap of two balls is inherently connected to the cost of transportation moving one ball to the other, yielding the Ollivier deﬁnition of Ricci curvature which is proposed by \cite{Ollivier_2009}.
\begin{myDef}[Olliver-Ricci Curvature]
In a graph, given the mass distribution $m_u^{\alpha}(v)$, node $\textbf{u} \in V$

% In a graph, given the mass distribution $m^{\alpha}_u(v)$ node u $\in V$
% \begin{equation}
%    [L_{ij}^\alpha] = \begin{cases} 
% \alpha, & \text{if } i = j, \\
% (1-\alpha) \frac{1}{D_{ii}}, & [A]_{ij} = 1, \\
% 0, & \text{otherwise}.
% \end{cases}
% \end{equation}
\begin{equation}\label{eq:mass_dist}
m^{\alpha}_u(v) = \begin{cases}
    \alpha, &  v = u, \\
    (1-\alpha) \frac{1}{|N(u)|}, & v \in N(u),\\
    0, & \text{otherwise}. 
\end{cases}
\end{equation}
\end{myDef}

Where $|N(u)|$ is the set of nodes linking to $u$.
The Ollivier’s Ricci curvature of a node pair $(i,j)$ is defined as follows:

\begin{equation}\label{eq:ricci_curvature}
    Ric^{\alpha}(i,j) = 1 - \frac{W({m_i^{\alpha},m_j^{\alpha}})}{d(i,j)}, 
\end{equation}

where $W({m_i^{\alpha},m_j^{\alpha}})$ is the Wasserstein distance between the probability (mass) distributions on nodes $i$ and $j$, and $d(i, j)$ is the distance deﬁned on the graph.
% \subsection{Graph Information Bottleneck}
% \begin{myDef}[Graph Information Bottleneck]
% For a graph $G$ and its label $Y$, the \textbf{Graph Information Bottleneck} aims to seek for the most informative yet compressed representation $Z$ following the Markov chain $<Y\to G \to Z>$:
% \begin{equation}
%     Z = \arg\min_{Z}-I(Z, Y)+\beta I(Z, G),
% \end{equation}
% where $\beta\ge 0$ is the Lagrangian multiplier trading off sufficiency and minimality and $I(\cdot,\cdot)$ is the mutual information. 
% \end{myDef}
% \begin{equation}
%     $\mathcal{L}=\mathcal{L}_{\rm CE}(Z, Y)+\beta \mathcal{D}_{\rm KL}\left (p \left (Z|G \right)||r\left (Z\right )\right )$; 
% \end{equation}

\begin{figure*}[t] % 'htbp' options for positioning
    \centering
    \includegraphics[width=0.95\textwidth]{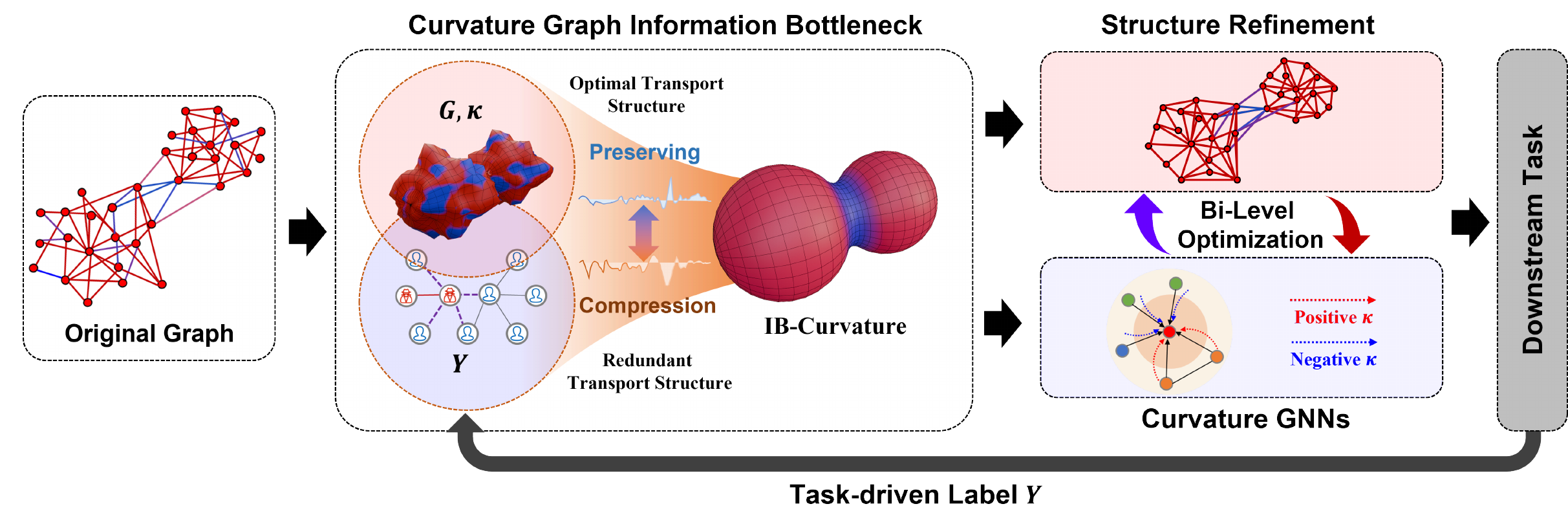}
    \vspace{-1em}
    \caption{The proposed \modelname~principle and its overall framework. }
    \label{fig:arch}
    \vspace{-1em}
\end{figure*}
\section{Discrete Curvature Graph Information Bottleneck}
We propose a supervised optimization model of graph structure based on \textit{Ricci flow} and \textit{VIB}, named as \modelname.
Specifically, we use \textit{Ricci flow} to optimize the discrete Ollivier-Ricci curvature guide by \textit{VIB}, learning a mapping function of curvature that measures the importance of all message edges in the neighborhood of the target node. 
% Aggregating information during message passing helps to classify information while reducing aggregation noise. 
% Finally, combined with the graph information bottleneck, the node representation is further compressed to improve the processing ability of the model for downstream tasks. 
The overall architecture of \modelname~ is shown in Fig.~\ref{fig:arch}.

\subsection{CurvGIB Principle Derivation}\label{sec:CurvGIB}
In this work, we focus on learning the critical information transport patterns during GNNs training, rather than directly reconstructing or learning the graph structures. 
Intuitively, we want to use the IB principle to find the "important" information in the training of GNNs, and then learn an optimal transfer metric for the critical information, named \textit{IB-Curvature}. 

Given a graph $G=(X,A)$, the Ricci curvature $\mathcal{K}$ of each edge can be obtained by Eq.~\ref{eq:ricci_curvature}. 
However, according to the definition of Eq.~\ref{eq:ricci_curvature}, we compute each time the Ricci curvature of the edge depends on the Wasserstein distance, which leads to unacceptable computational complexity. 
% The definition of Ollivier is based on optimal transport theory, which defines curvature as the change in the neighborhood mass distribution between nodes. 
Inspired by DeepRcci~\cite{sun2023deepricci}, we define the IB-guided optimization of the differentiable curvature following their approximate definition. 
Specifically, the mass distribution of the nodes is redefined using the Laplacian matrix $\boldsymbol{L}^\alpha = \alpha \boldsymbol{I} + (1-\alpha)\boldsymbol{D}^{-1}\boldsymbol{A}$ of the graph structure $A$. We can rewrite Eq.~(\ref{eq:mass_dist}) as: 
\begin{equation}
    \left[\boldsymbol{L}^{\alpha}\right]_{i j}=\left\{\begin{array}{ll}
\alpha, & i=j \\
(1-\alpha) \frac{1}{\boldsymbol{D}_{i i}}, & {[\boldsymbol{A}]_{i j}=1} \\
0, & \text { Otherwise }
\end{array}\right.
\end{equation}
where $\alpha$ is the mass weight of each node, and the $D$ is the diagonal degree matrix. 
Then, the differentiable IB-Curvature based on edge weight matrix $\mathcal{K}$ can be define as: 
\begin{equation}
\label{eq:diff_curv}
\begin{aligned}
  \kappa_{\rm IB}(i,j) = 1 - \frac{[\boldsymbol{L}^{\alpha} (\boldsymbol{A}) f(\mathbf{Z})]_{i} - [\boldsymbol{L}^{\alpha}(\boldsymbol{A}) f(\mathbf{Z})]_{j}}{d(\mathbf{z}_i, \mathbf{z}_j)},   
\end{aligned}
l\end{equation}
with an affine transform $f: \mathbb{R}^{N \times H} \rightarrow \mathbb{R}^{N}$ and the $d(\cdot,\cdot)$ is the geodesic distance in latent space.
According to Eq.~(\ref{eq:diff_curv}), it can be observed that this approximate curvature definition is differentiable with respect to representation $\mathbf{Z}$. 

We are concerned with the optimal transport structure of critical information,  
% Therefore, we further fixed the graph structure $A$ and used the transport weight matrix $A^{*}$ to represent the selection preference of important information on the message-passing path. 
and our curvature optimization relies only on node representation updates during training, which is more beneficial to using IB to guide the curvature update. 
Then we can give a formal definition of the \textit{Discrete Curvature Graph Information Bottleneck} (CurvGIB) for graphs based on \textit{IB-Curvature} by: 

\begin{myDef}[Curvature Graph Information Bottleneck]
For a graph $G=(X,A)$, label $Y$ and its Ricci curvature $\kappa$ of each edges, the optimal curvature $\kappa_{\rm IB}$ of node pair $(i,j)$ found by Information Bottleneck is denoted as:
\begin{equation}
\label{eq:CurvGIB}
\begin{aligned}
    \mathbf{Z}_{\rm IB}, \kappa_{\rm IB} &= \arg\min_{\mathbf{Z}, \kappa} \mathrm{CurvGIB}(\mathbf{X}, \mathbf{Y}, \mathbf{Z}, \kappa ) \\
    & \triangleq \arg\min_{\mathbf{Z}, \kappa} [ -I(\mathbf{Z} \mid \kappa ; \mathbf{Y})+\beta I(\mathbf{Z} \mid \kappa ; \mathbf{X}) ],
\end{aligned}
\end{equation}
% where the $d(\cdot,\cdot)$ is the geodesic distance in latent space.
\end{myDef}

% \begin{equation}I(Z|k;X)\leq\int p(x)p(y|(x,k))log\frac{P(y|(x,k))}{r(z|k)}dkdxdz\end{equation}

% where q,r is the variational posterior distribution.

Intuitively, the IB-curvature is updated with the hidden states learned by the IB-guided GNNs model and thus can be formulated as a constrained optimization form. 
For the IB constrain, the first term $-I(\mathbf{Z} | \mathcal{K}; Y)$ encourages that optimal transport information of label is preserved, and the second term $I(\mathbf{Z} | \mathcal{K}; \mathbf{X}))$ encourages that label-irrelevant information in message-passing is dropped. 
The learning procedure can be expressed through the Markov Chain: $\mathrm{MC}_{\mathrm{CurvGIB}}: < \mathbf{Y} \to \mathbf{Z}_{\kappa} \to \mathcal{K}>$. 
IB-Curvature represents the optimal transport pattern of the information most relevant to $Y$. 
As shown in Figure~\ref{fig:flows}, negative IB-Curvature represents that this critical information tends to be transported on the ``backbone" path of the graph, while positive IB-Curvature represents that critical information tends to propagate in the local highly connective structure.

\subsubsection{Variational Bounds of CurvGIB. }
We respectively introduce the lower bound of $I(\mathbf{Z}|\kappa ; \mathbf{Y})$ with respect to~\cite{poole2019variational}, and the upper bound of $I(\mathbf{Z}|\kappa ; \mathbf{X})$, for Eq.~\eqref{eq:CurvGIB}.

\begin{myPro}[\textbf{Lower bound of} $I(\mathbf{Z}|\kappa ; \mathbf{Y})$]
\label{prop:lower_bound}
For node representation $\mathbf{Z}$ with label $Y\in\mathbb{Y}$ and Ricci Curvature $\kappa$ learned from $A$, we have
\begin{equation}
\label{eq:lower_bound}
\begin{aligned}
    \!I(\mathbf{Z}|\kappa ; \mathbf{Y})\!\! \ge \!\!\int\! p((\mathbf{Z},\!\mathbf{Y}) | \kappa)\log\! q(\mathbf{Y} | (\kappa,\!\mathbf{Z}))d\mathbf{Y} d \kappa d\mathbf{Z}, 
\end{aligned}
\end{equation}
where $q$ is the variational posterior distribution.  
\end{myPro}

\begin{myPro}[\textbf{Upper bound of} $I(\mathbf{Z}|\kappa ; \mathbf{X})$]
\label{prop:upper_bound}
For node representation $\mathbf{Z}$ with label $Y\in\mathbb{Y}$ and Ricci Curvature $\kappa$ learned from $A$, we have
\begin{equation}
\label{eq:upper_bound}
\begin{aligned}
    I(\!\mathbf{Z}\!|\kappa ;\! \mathbf{X}) \! \!  \leq \! \! \! \int \! \!\! p(\mathbf{X})p(\mathbf{Y}|(\mathbf{X} \! , \! \kappa))\! \log \!\frac{p(\!\mathbf{Y}\!|(\mathbf{X}\!,\!\kappa))}{r(\mathbf{Z}|\kappa)} \! d\kappa d\mathbf{X} d\mathbf{Z},
\end{aligned}
\end{equation}
where $r$ is the variational posterior distribution.  
\end{myPro}
Proofs for Proposition~\ref{prop:lower_bound} and~\ref{prop:upper_bound} are provided in Appendix Proof section.

\subsection{Instantiating the \modelname~Framework}
% Following the above IB-Curvature definition, it can be observed that 
For simultaneously optimizing the IB curvature and learning the IB representation, we propose a bi-level optimization framework based on Ricci flow and curvature-aware GNNs.
The \modelname~framework instantiating as shown in Figure~\ref{fig:arch}.
% provides an implementation to optimize the discrete graph curvature to refine an optimal message-passing structure for GNNs alternately. 

\subsubsection{IB-Curvature and Representation Learning.}
Discrete graph curvature measures how easily the label information flows through an edge and can be guided to control the strength of message-passing. 
We first update the Ricci curvature of the refined structure $\boldsymbol{A}^{\star}$ according to Eq. 3 for each edge,
\begin{equation}
\begin{aligned}
    \label{Eq:ricci_ib}
        \kappa_{\rm IB}(i,j) = 1 - \frac{[\boldsymbol{L}^{\alpha} (\boldsymbol{A}^{\star}) f(\mathbf{Z})]_{i} - [\boldsymbol{L}^{\alpha}(\boldsymbol{A}^{\star}) f(\mathbf{Z})]_{j}}{d(\mathbf{z}_i, \mathbf{z}_j)}.
\end{aligned}
\end{equation}
Then, we advance~\cite{Ye2020CurvatureGN} by using an affine transform  $f: \mathbb{R}^{N \times M} \rightarrow \mathbb{R}^{N}$, and $f (\textbf{X}) = \textbf{W}\textbf{X} + \textbf{b}$. 
The loss and aggregation function of curvature-aware GNNs with curvature $\kappa$ are formulated as follows:
\begin{equation}
\begin{aligned}
    \label{Eq:GNN}
    \mathbf{Z}_{\kappa}&=\mathrm{CurvGNN}(\textbf{X}, \textbf{Y}, \kappa_{\rm IB}) \\
    \textsc{Agg}(\mathbf{Z}^{(l)}, \kappa_{\rm IB})&=\sigma\left(\mathbf{Z}^{(l)}+ f \left( \mathcal{K}^{(l)}(\kappa_{\rm IB}) \cdot \mathbf{Z}^{(l)}_{j} \right)\right),
\end{aligned}
\end{equation}
where $\mathcal{K}$ is the edge weight matrix obtained by IB-Curvature (we will introduce the definition later), and $\sigma$ is the activate function. 

Then, the IB principle guides GNNs to obtain critical information.
The optimal transport information representation $\mathbf{Z}_{\kappa}$ with fixed curvature $\kappa$ is:
\begin{equation}
\begin{aligned}
\mathbf{Z}_{\kappa} &= \arg\min_{\mathbf{Z}_{\kappa}} \mathrm{CurvGIB} \left( \mathbf{X}, \textbf{Y}, \mathbf{Z}, \kappa \right),\\
&= \arg\min_{\mathbf{Z}_{\kappa}} \left( -I(\mathbf{Z}_{\kappa}; Y) + \beta I(\mathbf{Z}_{\kappa}; \mathbf{X}) \right),
\end{aligned}
\end{equation}
where $\beta$ is Lagrangian multiplier, where smaller $\beta$ indicates more critical information was retained to $\mathbf{Z}_{\kappa}$.

\subsubsection{Structure Refinement with IB-Curvature. }
In differential geometry, Ricci flow evolves a smooth manifold, allowing positive and negative curvatures to evolve in their respective curvature directions, ultimately partitioning the entire manifold into multiple weakly connected sub-manifolds. 
Existing work primarily employs reverse Ricci flow to broaden the topological ``bottlenecks" of message passing, thereby alleviating issues such as over-squashing. 
In our paper, we aim to learn the optimal transport IB-curvature by using Ricci flow adaptively. 
Specifically, we follow existing work~ define a discrete Ricci flow as a edge-weighted graph with the edge weight matrix $\mathcal{K}$ for IB-Curvature as: 
\begin{equation}
\label{eq:Ricciflow}
\begin{aligned}
    \mathcal{K}^{(l+1)}=\left(1-\kappa^{(l)}_{\rm IB}\right) d^{(l)}(\mathbf{Z}_{\kappa}, \mathbf{Z}^T_{\kappa} ),
\end{aligned}
\end{equation}
where $l$ is the epoch number. 
According to Eq.~\ref{eq:diff_curv}, the updated Ricci curvature is: 
$1-\frac{[L^{\alpha}(A)f(Z_{\kappa})]_i-[L^{\alpha}(A)f(Z_{\kappa})]_j}{d(z_i^{\kappa},z_j^{\kappa})}$

For each pair of nodes, we jointly optimize the edge sampling probability $\pi_{ij}={\rm sigmoid}\left(\mathcal{K}_{ij}\right)$ by IB-Curvature weights and graph representation learning. 
We follow \cite{Jang2016CategoricalRW} to give the concrete relaxation of the Bernoulli distribution to update $\pi$:
\begin{equation}
\label{eq:Ricciupdate}
\begin{aligned}
    &[\boldsymbol{A}^{\star}]_{ij} \sim {\rm Ber}(\pi_{ij})\\
    &\approx {\rm sigmoid}\left(\frac{1}{\tau}\left(\log \frac{\pi_{u,v}}{1-\pi_{u,v}} + \log \frac{\epsilon}{1-\epsilon}\right)\right),
\end{aligned}
\end{equation}
where $\epsilon \sim {\rm  Uniform}(0,1)$ and $\tau\in \mathbb{R}^+$ is the temperature. 

The refined structure $A^{*}$ is optimized by maximizing the likelihood as
\begin{equation}
\label{eq:refine_struct}
\begin{aligned}
\boldsymbol{A}^{\star} = \arg \min_{\boldsymbol{A}^{\star}} \left [ -\sum_{i=1}^{N} \sum_{j=1}^{N} \log p\left([\boldsymbol{A}]_{i j}=1 \mid \boldsymbol{A}^{\star}\right) \right ]. 
\end{aligned}
\end{equation}

\subsubsection{Bi-level Optimization. }
According to Eq.~(\ref{eq:CurvGIB}) of the CurvGIB definition, the optimization of IB-Curvature $\kappa_{\kappa}$ and representation $\textbf{Z}_{\rm IB}$ depend on each other, we introduce a bi-level optimization mechanism~ to ensure the effectiveness and stability of training. 
The overall optimization function is formalized as follows:
\begin{equation}
\label{eq:bi-level}
\begin{aligned}
&\min \mathrm{CurvGIB} \left(\textbf{X},\textbf{Y}, \mathbf{Z}_\kappa, \kappa_{\rm IB} \right),\\
\mathbf{s.t.}~& \kappa_{\rm IB} \in \arg \min_{\kappa_{\rm IB}} \mathrm{IBCurv}(\mathbf{Z}_\kappa).\\
~& \mathrm{IBCurv}(\mathbf{Z}_\kappa) = \sum_{ij}(1-\kappa_{\rm IB}(i,j)) \dot d(\mathbf{z}_i, \mathbf{z}_j) 
\end{aligned}
\end{equation}

where $\mathrm{IBCurv}$ is obtained by combining Eq.~(\ref{Eq:ricci_ib}), (\ref{eq:Ricciflow}) and (\ref{eq:diff_curv}). 
The overall process of \modelname~is shown in Algorithm~\ref{alg:algorithm}.

\begin{algorithm}[!t]
\caption{The overall process of \modelname}
\label{alg:algorithm}
\LinesNumbered
\KwIn{Graph $G=(X, A)$ with label $Y$; Number of training epochs $E,E1,E2$;}
\KwOut{weighted node representation $Z$, predicted label $\hat{Y}$}
% \KwOut{IB-graph $G_{\rm IB}$, predicted label $\hat{Y}$}
% Let $\phi \leftarrow \phi_0$, $\theta \leftarrow \theta_0$\\
Parameter initialization;\\
% \tcp{Train \modelname}
% Initialization mass cost ${L\_cost}$ with Eq. (6);\\
% Initialization topology cost $Topo\_cost$ with Eq. (7);\\
% Initialize the edge weights $W$ with Eq. (2)(6)(7);\\
Initialization $\kappa$ with Eq.(10)

\For{$i=0,1,2,\cdots,E$}{
\tcp{IB-Curvature and Representation Learning}
\For{$i=0,1,2,\cdots,E1$}{
Aggregate node information with Eq.(11)
Optimize the first phase with Eq.(12)
}
% $X^{i+1},Z^{i} \leftarrow MLP(X^{0},W^{i})$;\\
% $X_{\rm IB} \leftarrow \{X_i \odot M, i\in |V|\}$;\\
% \tcp*{Feature masking}
% $A_{\rm IB} \leftarrow \bigcup_{u,v \in V}\{a_{u,v}\sim {\rm  Ber}(\pi_{u,v})\}$;\\
% % \tcp*{Structure learning}
% $G_{\rm IB}  \leftarrow (X_{\rm IB}, A_{\rm IB})$;\\
% \tcp{Learn distribution}
% Encode $(f^{\mu}_{\phi}(G_{\rm IB}), f^{\Sigma}_{\phi}(G_{\rm IB}))$ by a GNN;\\
% \tcp{Sample graph representation}
% Reparameterize $Z_{\rm IB}=f^{\mu}_{\phi}(G_{\rm IB})+f^{\Sigma}_{\phi}(G_{\rm IB})\odot\varepsilon$;\\
% % \tcp*{Reparameterize}
\tcp{Structure Refinement with IB-Curvature}
\For{$i=0,1,2,\cdots,E2$}{
Calculate edge weight  matrix $\mathcal{K}$  with Eq.(13)
 Sample structures with Bernoulli distribution by using $\mathcal{K}$ as a probability matrix with Eq.(14)
 
 Optimize the second phase with Eq.(15)
}
Optimize the overall framework with Eq.(16)
% $\mathcal{L}=\mathcal{L}_{\rm CE}(Z^{i}, Y)+\beta \mathcal{D}_{\rm KL}\left (p \left (Z^{i+1}|G \right)||r\left (Z^{i}\right )\right )$; \\
% Update model parameters to minimize $\mathcal{L}$.\\
% \tcp{Update edge weight metric}
% update topology cost $Topo\_cost$ with Eq. (4);\\
% update the edge weights $W$ with Eq. (2);\\

}
\end{algorithm}

\begin{table}[t]
\small
\centering
\begin{tabular}{cl|rrrrr} 
\toprule
\multicolumn{2}{c|}{\textbf{Dataset}}  & \textbf{\#Nodes} & \textbf{\#Edges} &\textbf{Avg. Degree} & \textbf{\#Labels} \\
\midrule
&\textbf{Cora}     & 2,708  & 5,429 & 3.90  & 7        \\
&\textbf{Citeseer}  & 3,312 & 4,732 & 2.79  & 6        \\
&\textbf{PubMed}  &19,717   &44,338   &4.50    &3        \\
&\textbf{CS}  &18,333   &81,894   &8.93    &40        \\
&\textbf{Physics}  &34,493   &247,962   &14.38    &5        \\
&\textbf{Amazon-C}  &13,752   &287,209   &41.75    &10        \\
&\textbf{Amazon-P}  &13,381   &245,778   &36.71    &10        \\
\bottomrule
\end{tabular}
\caption{Statistics of datasets.}
\label{dataset_description}
\vspace{-1em}
\end{table}

\renewcommand{\arraystretch}{1.2}\label{tab:2}
\begin{table*}[!h]

% \begin{table*}[htbp]

\centering
\small
\resizebox{\textwidth}{!}{
\begin{tabular}{c|cc|cc|cc|cc|cc|cc|cc}
\toprule
\multirow{2}{*}{\textbf{Method}} & \multicolumn{2}{c}{\textbf{Cora}}    & \multicolumn{2}{c}{\textbf{CiteSeer}}     &\multicolumn{2}{c}{\textbf{PUBMED}}   & \multicolumn{2}{c}{\textbf{CS}}   & \multicolumn{2}{c}{\textbf{PHYSICS}} & \multicolumn{2}{c}{\textbf{AMAZON-C}}  & \multicolumn{2}{c}{\textbf{AMAZON-P}}
    \\ 
\cline{2-15} 

        & \small ACC\               &  \small Macro-F1            & \small ACC                  & \small Macro-F1  
          & \small ACC                  & \small Macro-F1 
          & \small ACC                  & \small Macro-F1 
          & \small ACC                  & \small Macro-F1 
          & \small ACC                  & \small Macro-F1 
          & \small ACC                & Macro-F1 \\ 
\midrule

GCN   &  81.5±0.5 &81.9±1.0
& 70.9±0.5 & \underline{82.4±0.9}
& 79.0±0.3 &\underline{ 81.8±0.5}
&  91.1±0.5 &82.5±0.6
& 92.8±1.0 &81.8±0.5
& 82.6±2.4 & 82.4±0.6
& 91.2±1.2 &81.4±0.4  \\

GAT    &   83.0±0.7  &76.8±17.8 
&  72.5±0.7 &   77.4±17.9
&  79.0±0.3 &72.6±23.4  
&  90.5±0.6  &75.8±18.9
&   92.5±0.9  &74.5±21.1  
&  78.0±19.0 &75.9±21.4
&  85.1±20.3 &76.1±19.7         \\

GIN &    86.6±0.4  &77.6±8.9 
&   63.4±0.6    &77.9±8.5  
&   86.6±0.1 &76.8±7.7   
&   53.3±0.4  &78.5±8.2 
&    86.6±0.1  &75.7±7.7  
&   54.3±19.0 &77.7±8.9
&   51.4±17.0  &77.3±7.8    \\  

GraphSAGE &    83.4±3.2 & \underline{82.2±8.1}
&   73.4±4.9   &\textbf{84.7±8.5}
&   \underline{86.9±4.7}   &\textbf{86.1±5.5}
&   92.4±5.7 &84.0±8.2
&    \underline{95.0±4.8} &83.7±7.0 
&   80.4±13.7 &\textbf{84.2±7.3}
&   89.7±7.7 &\underline{85.1±6.4}\\ 
\midrule

SDRF &     86.3±0.3    &58.8±29.6
&    72.6±0.3    &59.5±13.7
&    85.2±0.1    &66.4±23.1
&    92.2±0.1    &65.4±25.5
&     OOM    &OOM  
&    \underline{86.6±0.2}    &60.0±24.9
&    91.6±0.3    &63.9±25.6       \\

BORF &    \underline{87.5±0.2} &62.7±26.7
&   \underline{73.8±0.2} &58.7±14.3 
&   86.1±0.4 &68.0±21.5 
&   \underline{93.1±0.1} &71.8±21.9
&    OOM &OOM  
&   OOM &OOM
&   \underline{93.6±0.2} &66.6±24.6      \\ 

 % CurvGN-n &  82.7±0.7 &79.1±0.5
 % & 72.1±0.6 &68.3±0.3
 % & 79.2±0.5 &75.8±0.7
 % & 92.9±0.3 &91.8±0.3
 % & 94.3±0.2 &92.5±0.3 
 % & 86.5±0.7 &79.5±1.2
 % & 92.5±0.5 &87.3±1.2       \\
  CurvGN-n &  82.7±0.7 &78.3±0.5
 & 72.1±0.6 &65.4±0.4 
 & 79.2±0.5 &75.4±0.8
 & 92.9±0.3 &86.4±0.8
 & 94.3±0.2 &85.2±0.6  
 & 86.5±0.7 &69.1±2.2
 & 92.5±0.5 &80.6±1.7       \\

% \midrule

\midrule

SUBLIME &  84.2±0.5 &71.7±0.5
 & 73.5±0.6 &66.4±1.4
 & 81.0±0.6 &73.7±1.6 
 & 92.3±0.7 &\underline{87.4±1.7}
 & 94.2±0.3 &\textbf{91.9±0.5 } 
 & 72.3±0.5 &42.0±4.0
 & 81.3±0.3 &66.6±2.2       \\

MVPOOL-SL &  83.6±0.4 &78.1±7.6
& 72.6±0.5 &61.1±3.2 
& 80.0±0.4 &60.1±11.2
& 92.0±0.5 &84.9±6.2 
& 94.0±0.6 &\underline{90.0±1.1} 
& 79.5±1.5 &68.5±16.9
& 88.5±1.2 &79.0±16.3 \\ 
\midrule
 \textbf{CurvGIB} &  \textbf{89.0±0.9} &\textbf{84.3±0.9}
 & \textbf{75.5±0.1} &73.5±1.4
 & \textbf{88.0±0.1} &78.9±0.9
 & \textbf{94.2±0.5} &\textbf{91.2±0.5}
 &  \textbf{96.5±0.3} &87.7±0.5 
 & \textbf{91.9±1.4} &\underline{83.9±1.7}
 & \textbf{93.7±1.4} &\textbf{94.8±4.1  }     \\
\bottomrule
\end{tabular}}
\centering
\caption{
Summary of node classification results: “average accuracy ± standard deviation”. 
\underline{Underlined}: best performance of specific backbones, \textbf{bold}: best results of each dataset.
}
\label{table:results}
\vspace{-1em}
\end{table*}

\begin{figure*}[htbp]
\centering
\subfigure[Parameter Sensitivity Analysis]{
a\includegraphics[width=1\columnwidth]{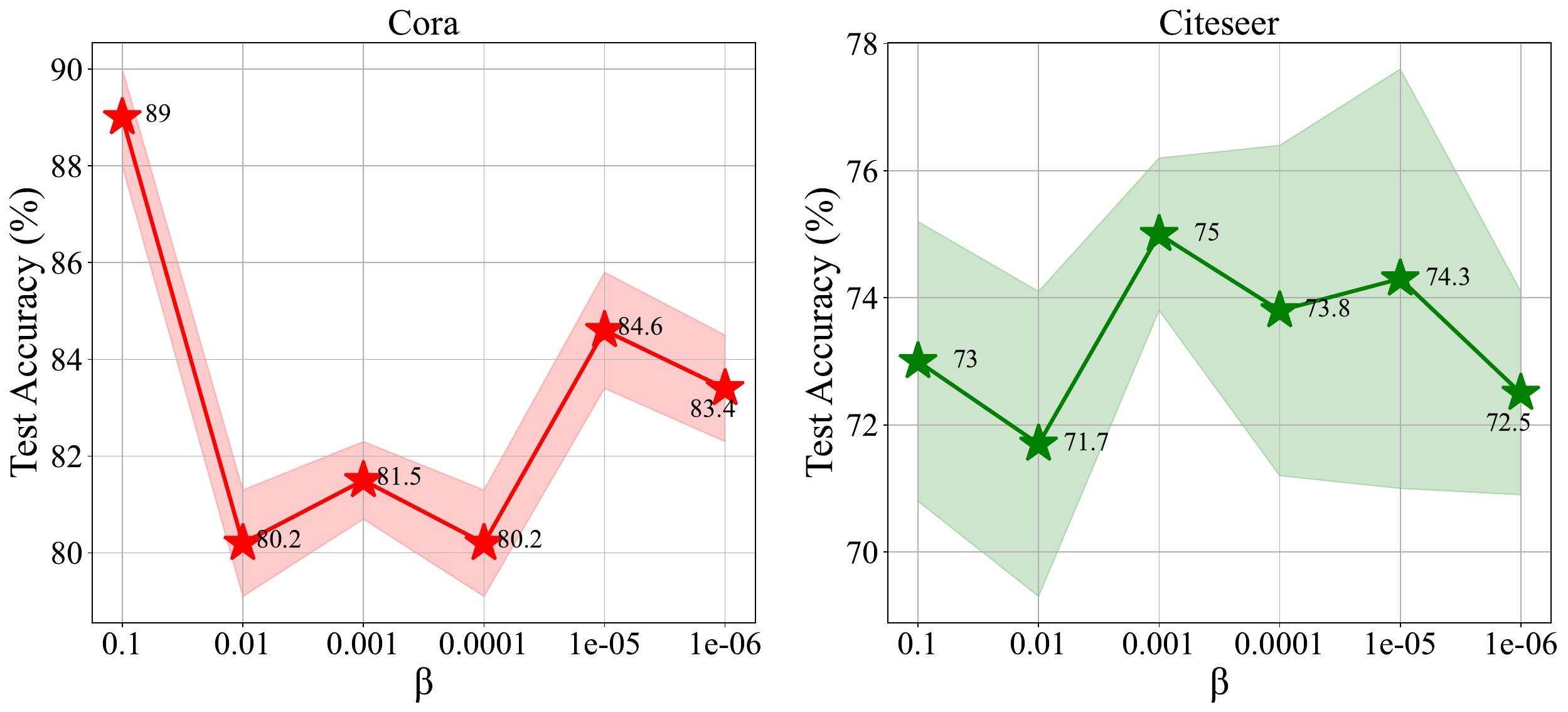} 
}
\subfigure[Performance with Different Ratio]{
\includegraphics[width=1\columnwidth]{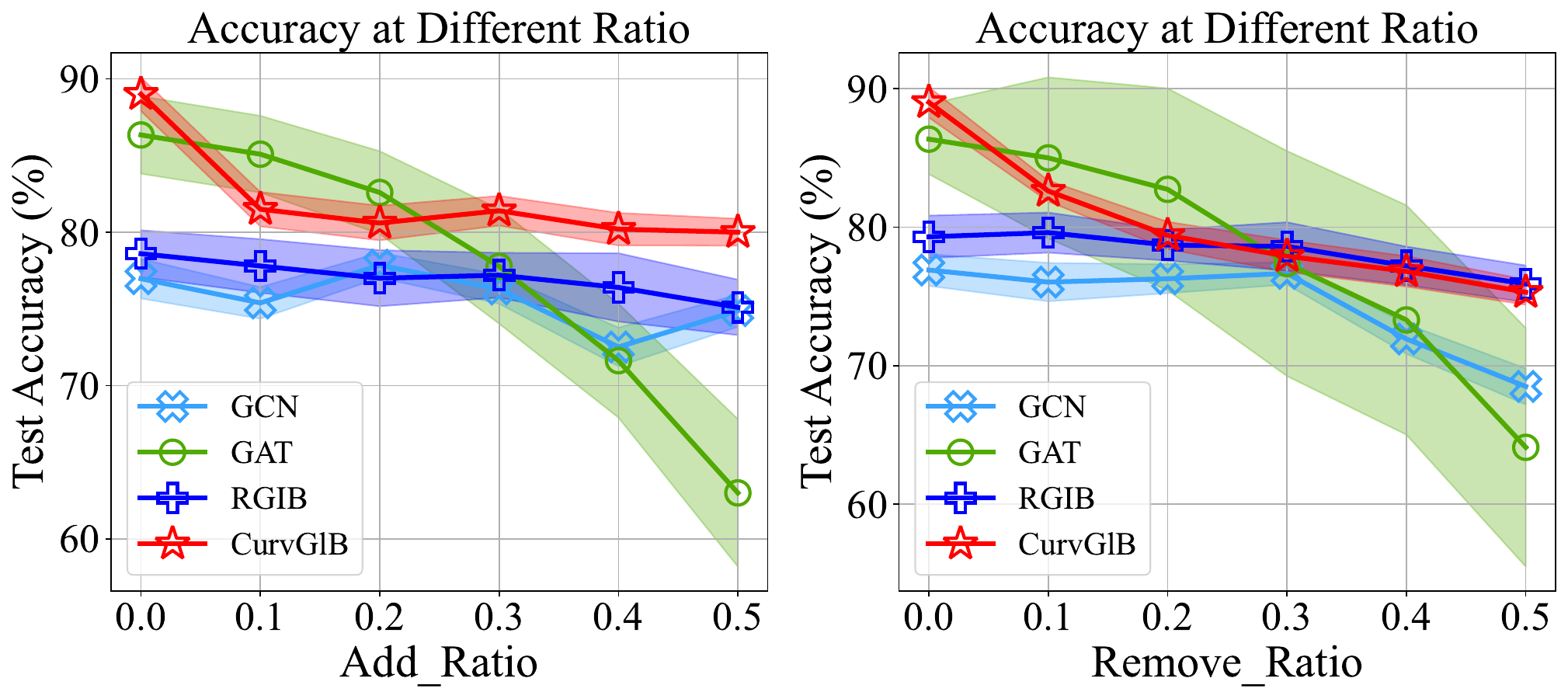} 
}
\vspace{-1em}
\caption{Parameter sensity and robustness analysis on Cora and Citeseer.}
\label{fig:analysis_synth}
\vspace{-1em}
\end{figure*}

\section{Experiments}
We evaluate \modelname~\footnote{\texttt{{https://github.com/RingBDStack/CurvGIB}}.} on two tasks: node classification and graph denoising, to verify whether \modelname~ can retain critical structures conducive to message passing to improve the efficiency and robustness of graph representation learning. 
% We then design experiments to verify the stability of \modelname~ and further demonstrate that \modelname~ can compress unnecessary redundant information while preserving the critical graph structures by visualizing both synthetic and realworld datasets which are processed by \modelname~ and baselines .
We then perform an extensive and comprehensive validation analysis to further demonstrate that \modelname~ can preserve critical graph structure while compressing unnecessary redundant information.

\underline{}

\subsection{Datasets}
As shown in Table \ref{dataset_description}, we conduct experiments on several real-world datasets: 
(1) Citation network: Cora and Citeseer~\cite{kipf2017semisupervised} are citation networks of machine learning academic papers and PubMed is a citation network of biomedical academic papers.
(2) Co-occurrence network~\cite{shchur2019pitfallsgraphneuralnetwork}: 
Coauthor CS and Coauthor Physics are co-authorship graphs based on the Microsoft Academic Graph from the KDD Cup 2016 challenge.
Amazon Computers and Amazon Photos~\cite{shchur2019pitfallsgraphneuralnetwork} which are segments of the Amazon co-purchase graph.

\subsection{Experimental Setup}
% \begin{figure}[htbp]
%     \centering
%     \includegraphics[width=1\linewidth]{photos_paper/combined_plot.pdf}
%     \caption{Model robustness under different num layers}
%     \label{fig:enter-label}
% \end{figure}

\begin{figure*}[ht]
\centering
\subfigure[Cora ]{
\includegraphics[width=0.33\linewidth]{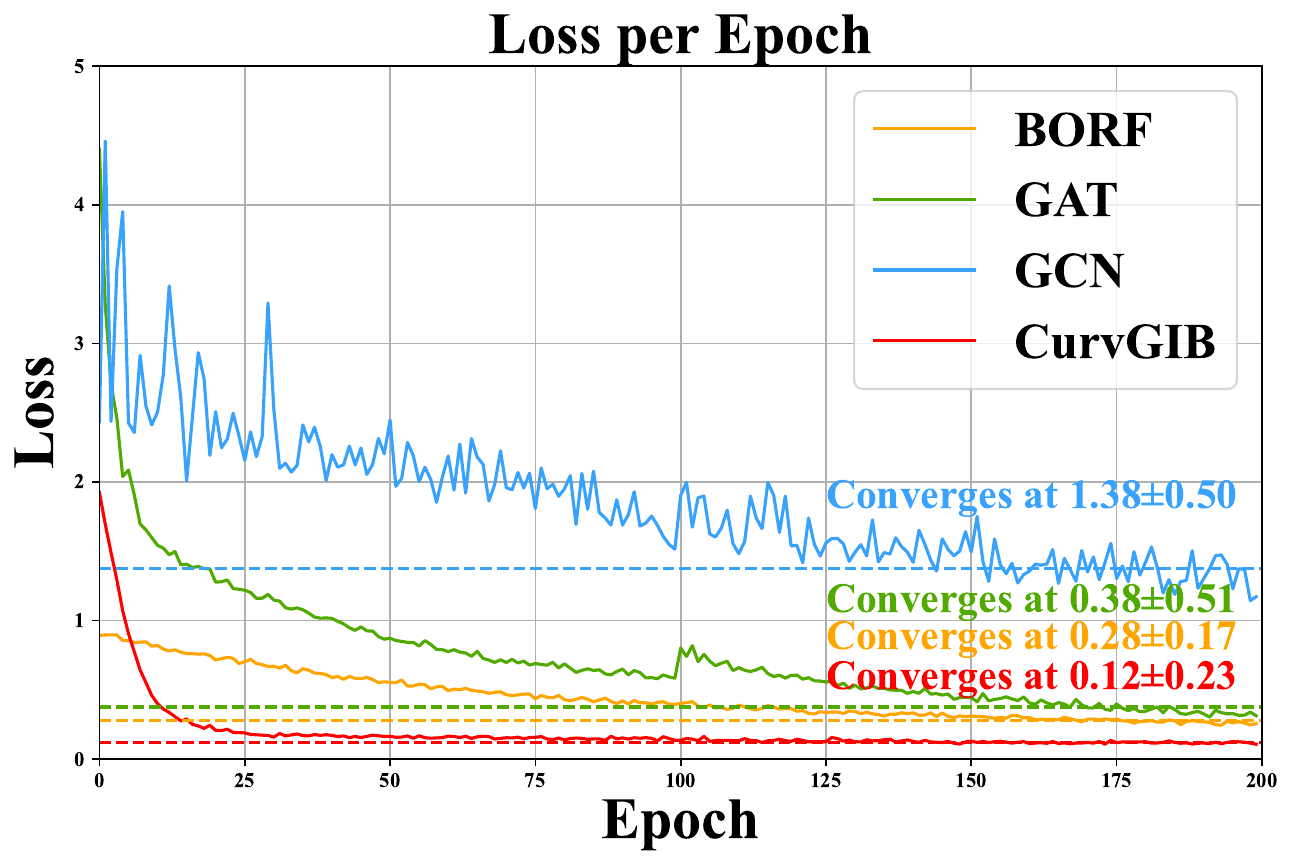}
%\caption{fig1}
}%
\subfigure[CiteSeer]{
\includegraphics[width=0.33\linewidth]{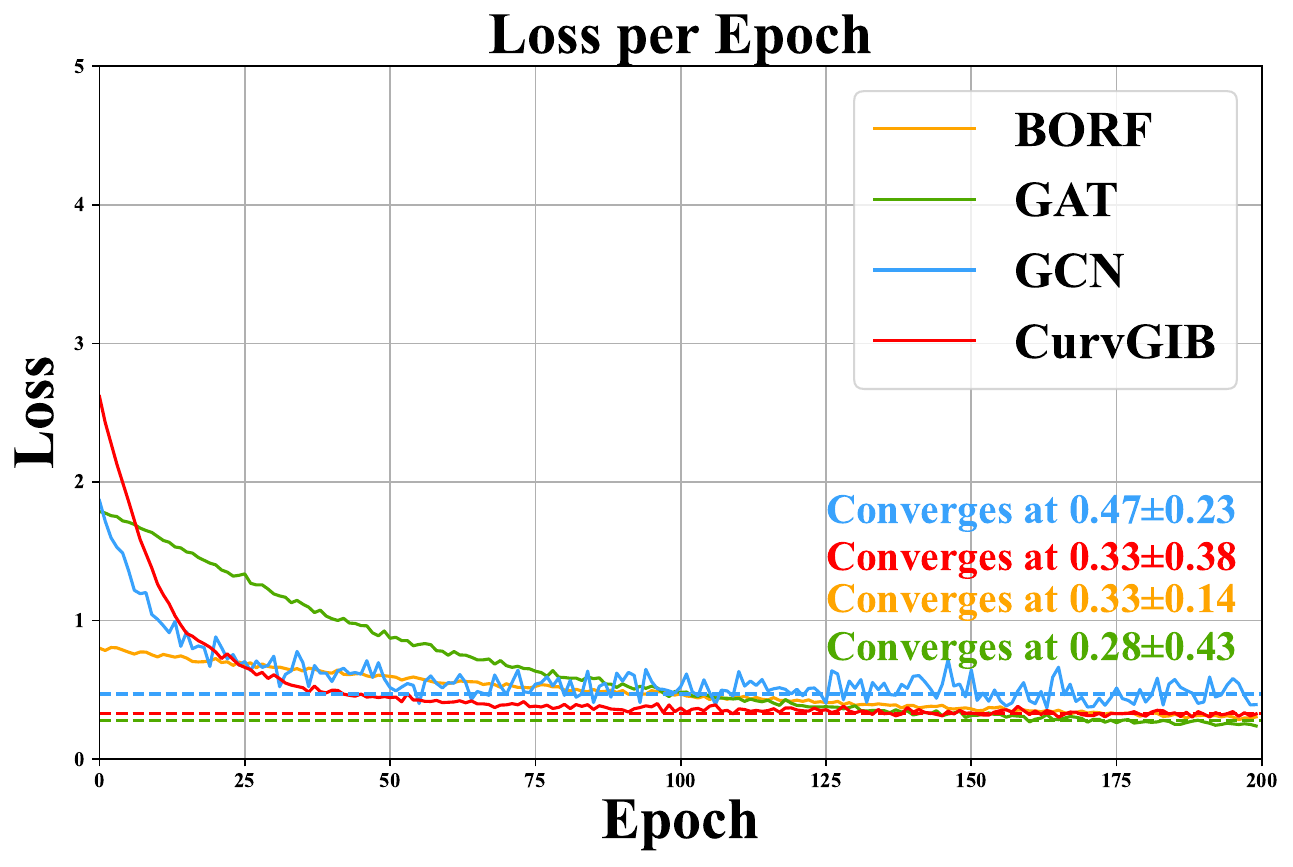}
%\caption{fig2}
}%
\subfigure[Pubmed ]{
\includegraphics[width=0.33\linewidth]{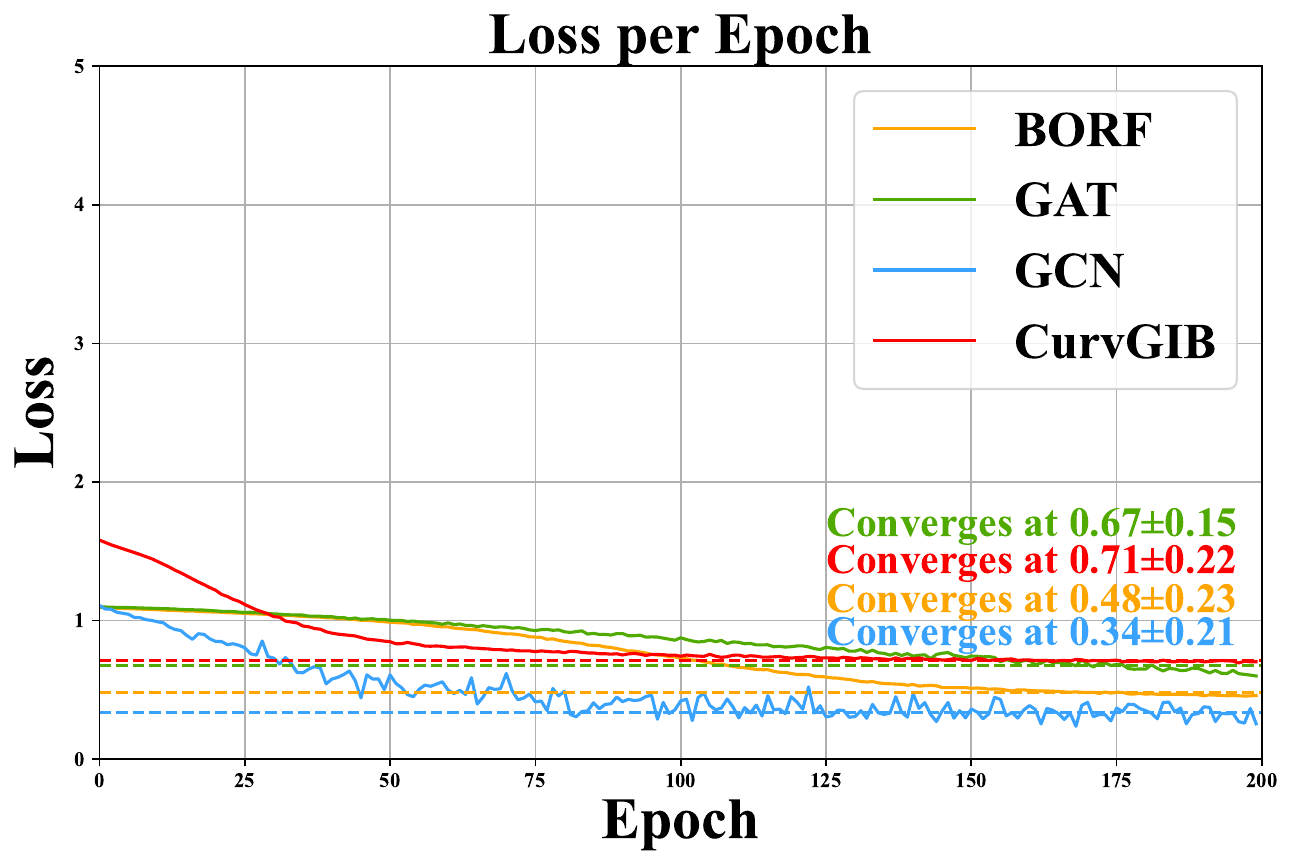}
%\caption{fig2}
}%
\centering
% \caption{({\textbf{blue solid lines}}) of a triangular structure approximate a tree (\textbf{1})}
\vspace{-1em}
\caption{Learning Stability Analysis on Different Datasets}

\label{fig:Learning}
\end{figure*}

\begin{figure*}[htbp]
\centering
\subfigure[Original]{
\includegraphics[width=0.24\linewidth]{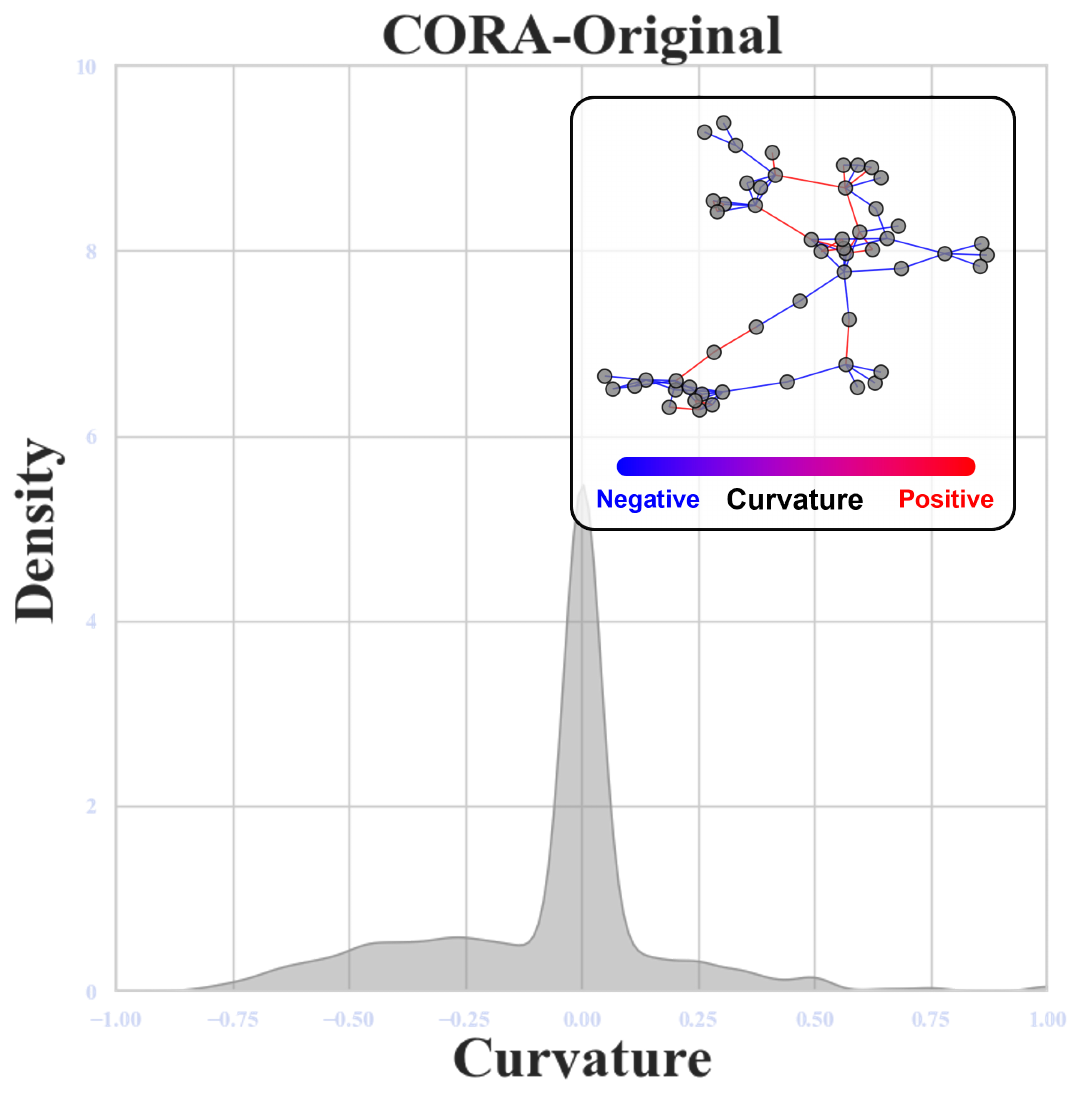}
%\caption{fig1}
}%
\subfigure[BORF]{
\includegraphics[width=0.24\linewidth]{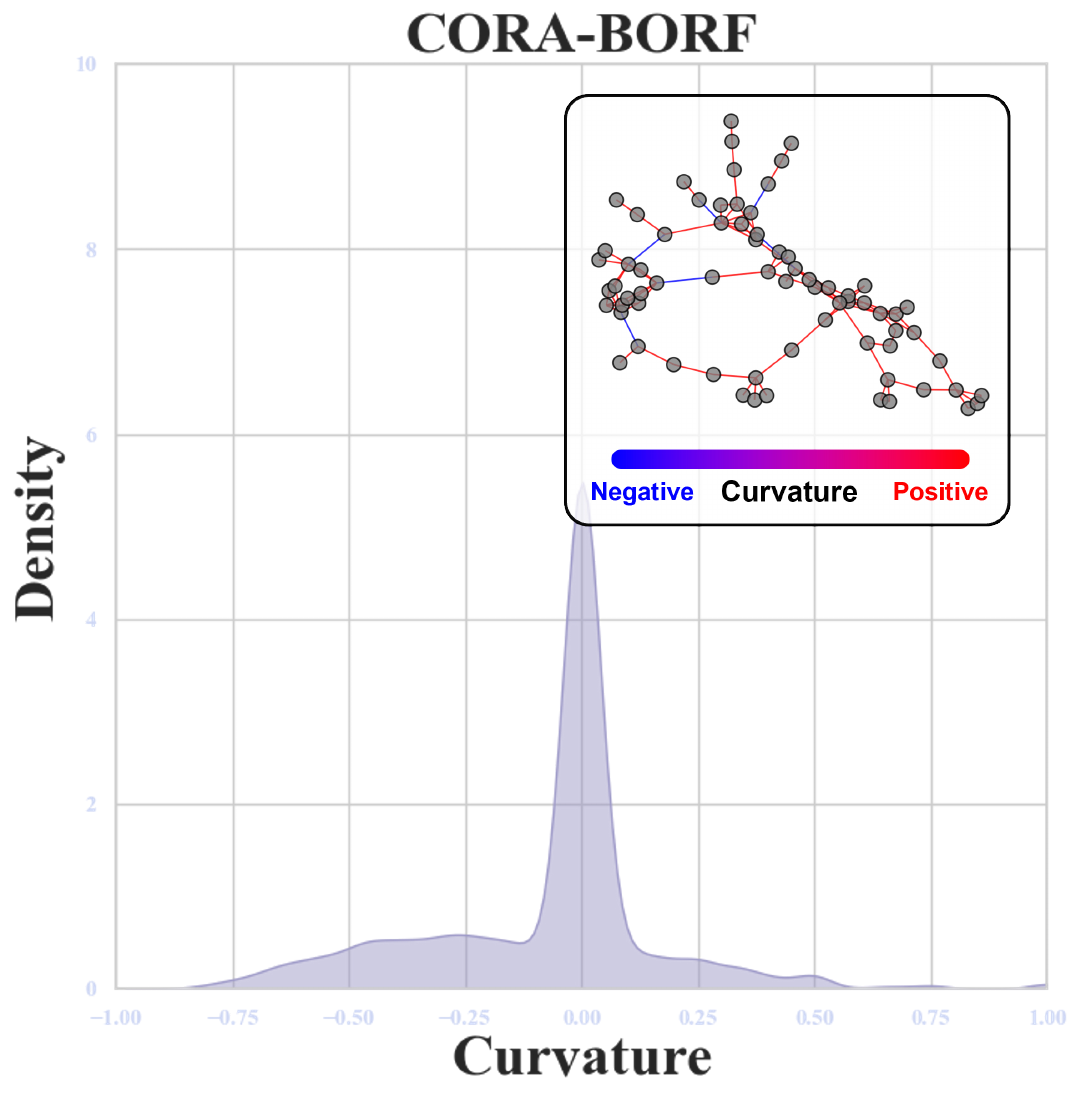}
%\caption{fig2}
}%
\subfigure[\textbf{CurvGIB(Ours)}]{
\includegraphics[width=0.24\linewidth]{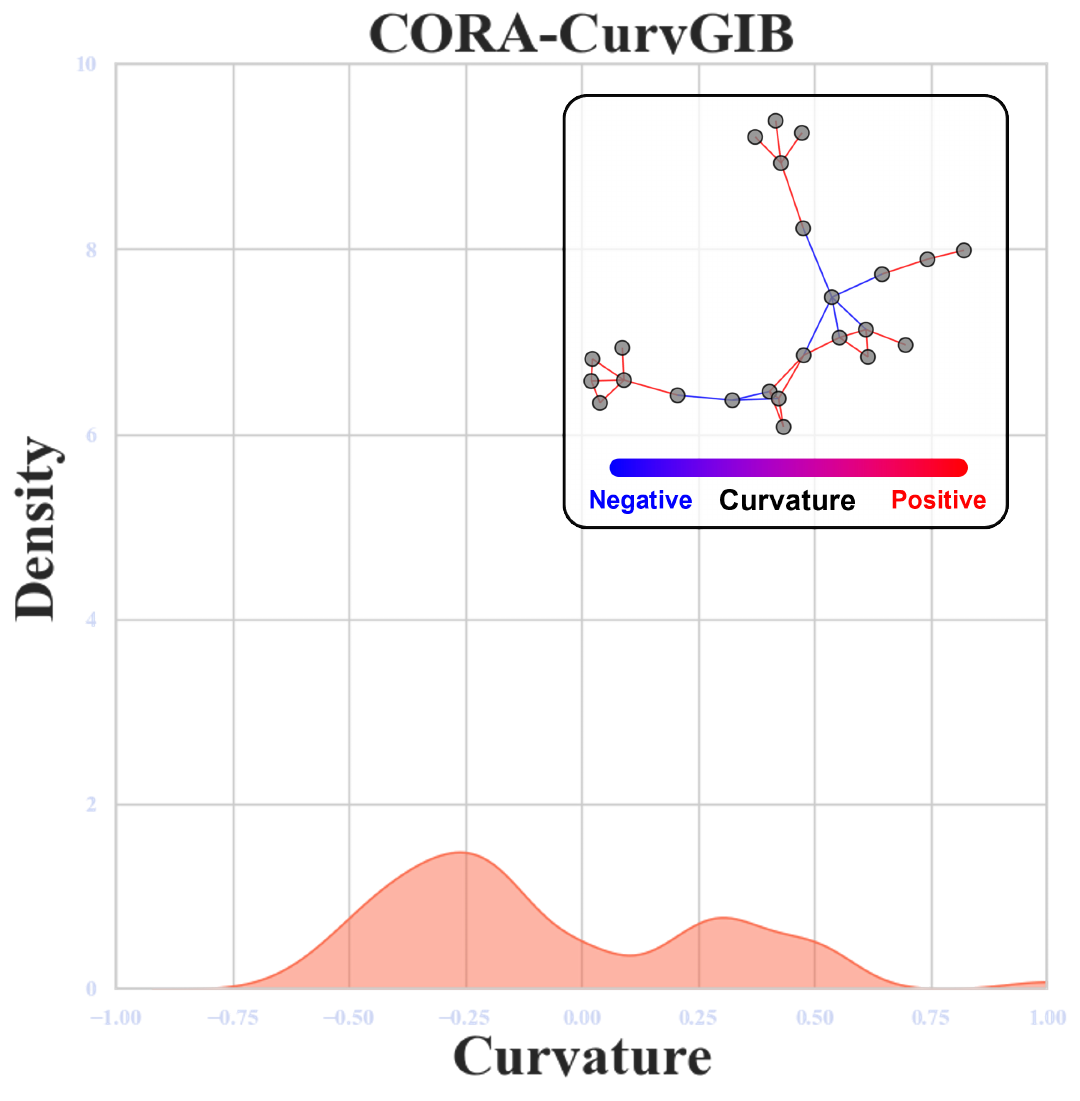}
%\caption{fig2}
}%
\subfigure[VIB-GSL]{
\includegraphics[width=0.24\linewidth]{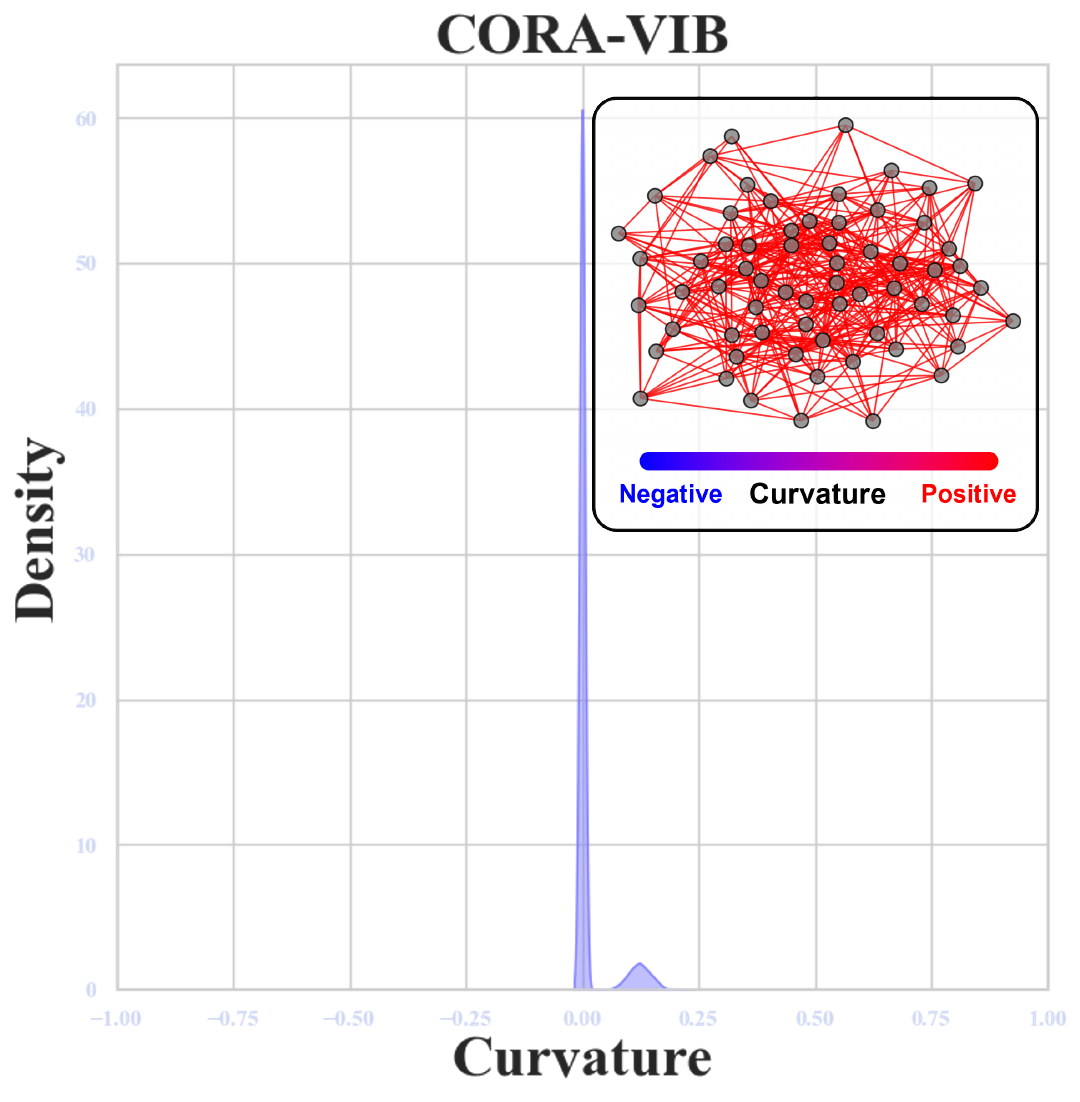}
%\caption{fig2}
}%
\vspace{-1em}
\caption{Distributions of Ricci curvature and visualizations of sampling sub-structures after learning on Cora.}
\label{fig:curvatures}
\vspace{-1em}
\end{figure*}

% To evaluate the proposed CurvGIB, we compare it with several representative baselines, including GCN, GAT, GIN, and GraphSAGE. 
% Additionally, we incorporate graph curvature method baselines such as SDRF,  BORF and CurvGN. 
% SDRF is an edge rewiring method designed to address the over-squashing issue by removing and adding edges based on Ricci curvature. 
% BORF is a curvature based rewiring method that can effectively improve GNN performance. 
% CurvGN is the first graph convolutional network based on graph curvature. CurvGN adapts to various local structures and filters messages exchanged between nodes in different ways.
% We also extend our evaluation to include node level graph structure learning baselines.
% SUBLIME optimizes structure by maximizing the alignment between the learned structure and a crafted, self-enhanced learning target using contrastive learning.
% MVPOOL-SL designs a structure learning mechanism for the pooled graph, enhancing its ability to learn a refined graph structure that effectively preserves the underlying substructures of the graph.\\
\subsubsection{Baselines.}
We compare the \modelname~ with a number of node-level baselines which can be divided into 4 categories, including traditional graph neural networks \cite{kipf2017semisupervised, Xu2018HowPA, Hamilton2017InductiveRL, Velickovic2017GraphAN}, graph structure learning method \cite{Liu2022TowardsUD, Zhang2021HierarchicalMG}, information bottleneck method\cite{Wang2022TowardER} and Graph curvature method\cite{Ye2020CurvatureGN,topping2021understanding, Nguyen2022RevisitingOA}, to demonstrate the effectiveness and robustness of \modelname~. 
% We do not include VIB-GSL in our baselines since it focuses on graph-level representation learning. We plug GCN \cite{Kipf2016SemiSupervisedCW} into \modelname~  to see whether the \modelname~ can boost the performance of node classification or not. 

% \begin{figure}
%     \centering
%     \includegraphics[width=1.0\linewidth]{photos_paper/canshu.pdf}
%     \caption{Caption}
%     \label{fig:enter-label}
% \end{figure}
% \begin{figure}
%     \centering
%     \includegraphics[width=01.0\linewidth]{photos_paper/Ratio.pdf}
%     \caption{Caption}
%     \label{fig:enter-label}
% \end{figure}
% \textbf{Graph Visualization.}

\subsubsection{Parameter Setting.}
We set both the information bottleneck size $K$ and the embedding dimension of baseline methods as 64. For CurvGIB,we perform the depth search of $l \in\{2,4,6,8\} $ for each dataset,and perform hyperparameter search of  $\beta \in \{10^{-1},10^{-2},10^{-3},10^{-4},10^{-5},10^{-6}\}$ for each dataset.

\subsection{Results And Analysis}
\subsubsection{Node Classification.}
We first examine CurvGIB's performance on node classification task. 
We perform 10-fold cross-validation and report the average accuracy, average F1 score, and the standard deviation across the 10 folds in Table \ref{table:results}. 
The suboptimal results of each dataset are underlined and the best results of each dataset are shown in bold. 
As shown in Table \ref{table:results}, the proposed CurvGIB consistently outperforms all baselines on all datasets. Especially in the
dataset Amazon-Computers, our method outperforms all baselines by a large margin. Generally, the graph structure learning method(SUBLIME, $\rm MVPOOL_{SL}$) only shows small improvement or even have a negative impact on some datasets (Amazon-Photos and Amazon-Computers), which is because they are constrained by the
generated graph structures which miss the critical structure information from the original graph. 
The graph structure optimized method based on Ricci curvature (BORF, SDRF, CurvGN, CurvGIB) all perform well. 
Our method gets an improvement over methods based on Ricci curvature.

% \textbf{Trade-off between Prediction and Compression.
% }
\subsubsection{Parameter Sensity.}
We explore the influence of the Lagrangian multiplier $\beta$ trading off prediction and compression in Eq.(\ref{Eq:IB}, \ref{eq:CurvGIB}, \ref{eq:lower_bound}, \ref{eq:upper_bound}). 
% Note that there is a relationship between increasing $\beta$ and decreasing $K$ \cite{Shamir2008LearningAG}, and the following analysis is with $K$ = 16.
Figure~\ref{fig:analysis_synth} (a) depicts the changing trend of node classification accuracy on Cora and Citeseer.
Based on the results, we make the following observations: (1) Remarkably,
the node classification accuracies of \modelname~ variation across different $\beta$ show the same changing behavior on both datasets. 
Since Citeseer is a concatenation of many different small graphs, a high $\beta$ results in the model not learning enough information to fit the features of the subgraphs. 
Instead, Cora requires a higher $\beta$ to compress more information. 
% Therefore, it can be seen that although the two datasets exhibit the same changing behavior, the overall shape of graph is different.
(2) Appropriate value of $\beta$ can greatly increase the model’s performance. \modelname~ achieves the best balance of prediction and compression with $\beta=10^{-1}$ and $\beta=10^{-3}$  on Cora and Citeseer, respectively.
This indicates that different datasets consist of different percent of task-irrelevant information and hence need a different degree of information compression.

\subsubsection{Graph Denoising.}
To evaluate the robustness of our framework by adding or deleting edges on Cora. 
Specifically, for each graph in the dataset, we randomly remove (if edges exist) or add (if no such edges) $10\%, 20\%, 30\%,40\%,50\%$ edges. 
As shown in Figure~\ref{fig:analysis_synth} (b), the classification accuracy of GCN dropped by 10\% with 50\% missing edges and dropped by $5\%$ with $50\%$ noisy edges, indicating that GNNs are indeed sensitive to structure noise. 
% Since \modelname~does not depend on the noisy original structure, instead it learns an edge weight coefficient based on Ricci curvature and 
\modelname~achieves better results with small performance degradation, and with the increase of the ratio, the accuracy of \modelname~ remains stable gradually. 
\subsubsection{Learned Graph Curvature.}
We tested different models on the Cora dataset and visualized the Ricci curvature of edges before and after processing. 
As shown in Figure \ref{fig:curvatures}, \textbf{BORF} tends to remove edges with negative curvature as much as possible while ensuring that the curvature distribution remains unchanged.
\textbf{VIB-GSL}~\cite{Sun2021GraphSL} tends to process the original graph into a fully connected graph.
Our model does not have special preferences compared to the above models.
Instead, it adjusts to the structure that is most conducive to message passing based on the specific dataset.

% \begin{figure}
%     \centering
%     \includegraphics[width=0.5\linewidth]{photos_paper/exp/STABLE_ACC.png}
%     \caption{Caption}
%     \label{fig:enter-label}
% \end{figure}
% \begin{figure}
%     \centering
%     \includegraphics[width=0.5\linewidth]{photos_paper/exp/w_distance.png}
%     \caption{Caption}
%     \label{fig:enter-label}
% \end{figure}
% \begin{figure}
%     \centering
%     \includegraphics[width=0.5\linewidth]{photos_paper/exp/loss_stable.png}
%     \caption{Caption}
%     \label{fig:enter-label}
% \end{figure}
\subsubsection{Training stability.}
\modelname~ deduces a tractable variational approximation for the \textbf{IB} objective, which facilitates the training stability.
In this subsection, we analyze the convergence of \modelname~ and other baselines, such as \textbf{BORF}\cite{Nguyen2022RevisitingOA}, \textbf{GCN}\cite{kipf2017semisupervised}, \textbf{GAT}\cite{Velickovic2017GraphAN}, on Cora, Citeseer and Pubmed with a learning rate of 0.001.
% Figure 5 (a) depicts the losses of \modelname~ (i.e., overall loss L, cross-entropy loss LCE for classification, and the KL-divergence loss DKL) with $\beta = 10^{-3}$ .
Figure~\ref{fig:Learning} shows the overall loss of \modelname~ and baselines. 
As shown in Figure~\ref{fig:Learning}, except \textbf{GCN}, other models all coverage steadily.
However, \modelname~ converges 2.5 times as fast as model \textbf{GAT} and converges 2 times as fast as model \textbf{BORF}.
It benefits from our \modelname principle, which improve the model to retain critical structures for sufficient information transport.
\section{Conclusion}

In this study, we introduced the Discrete \textbf{Curv}ature \textbf{G}raph \textbf{I}nformation \textbf{B}ottleneck (\textbf{\modelname}) framework, which demonstrates significant advantages in optimizing the information transport structure of Graph Neural Networks (GNNs). By incorporating Ricci curvature, \modelname~enhances information propagation efficiency in complex graph structures from a geometric perspective, providing a more precise and effective approach compared to direct rewiring or learning graph structures. The main contributions of \modelname~include optimizing discrete curvature to improve graph structures, extending the Information Bottleneck (IB) principle to learn optimal information transport patterns for specific downstream tasks, reducing computational complexity through a tractable curvature optimization approximation, and demonstrating superior performance and interpretability through extensive empirical validation. However, the current research focuses primarily on node-level tasks and has not yet been extended to graph-level tasks. Future work should explore how \modelname~can be adapted for graph-level tasks and improve its applicability and effectiveness in different scenarios and applications.

\section*{Acknowledgments}
The corresponding authors are Xingcheng Fu and Qingyun Sun. The authors of this paper are supported by the National Natural Science Foundation of China through grants No.U21A20474, No.62462007 and No.62302023, and the Research Fund of Guangxi Key Lab of Multi-source Information Mining \& Security (24-A-02-01). We extend our sincere thanks to all authors for their valuable efforts and contributions.

\bibliography{ref}
\clearpage
\appendix
\section{Proofs}

\subsection{Proof of Proposition~\ref{prop:lower_bound}}
We first provide the proof of proposition~\ref{prop:lower_bound} in Section~\ref{sec:CurvGIB}.

\begin{proof}
Our goal can be expressed as: 
\begin{equation}
\label{eq:curvib}
\begin{aligned}
\mathbf{Z}_{\mathrm{IB}},\kappa_{\mathrm{IB}}&=\arg\min_{\mathbf{Z},\kappa}\mathrm{CurvGIB}(\mathbf{X},\mathbf{Y},\mathbf{Z},\kappa)\\&\triangleq\arg\min_{\mathbf{Z},\kappa}[-I(\mathbf{Z} | \kappa;\mathbf{Y})+\beta I(\mathbf{Z} | \kappa;\mathbf{X})]. 
\end{aligned}
\end{equation}

According to the definition of mutual information, for the first part of the Eq~\eqref{eq:curvib}:

\begin{equation}
\begin{aligned}
\label{eq:curvib_first}
I(\mathbf{Z} | \kappa;\mathbf{Y})=&\iint p(\mathbf{Z} | \kappa;\mathbf{Y})\log\frac{p(\mathbf{Z} | \kappa;\mathbf{Y})}{p(\mathbf{Z} | \kappa)p(\mathbf{Y})}\mathrm{d}(\mathbf{Z} | \kappa)\mathrm{d}\mathbf{Y}
\\=&\iint p(\mathbf{Z} | \kappa;\mathbf{Y})\log\frac{p(\mathbf{Y} |  (\mathbf{Z} | \kappa))}{p(\mathbf{Y})}\mathrm{d}(\mathbf{Z} | \kappa)\mathrm{d}\mathbf{Y}
\\=&\iiint p((\mathbf{Z},\mathbf{Y}) | \kappa)\log\frac{p(\mathbf{Y} |  (\mathbf{Z},\kappa))}{p(\mathbf{Y})}\mathrm{d}\mathbf{Z}\mathrm{d}\kappa\mathrm{d}\mathbf{Y}
\\=&\iiint p((\mathbf{Z},\mathbf{Y}) | \kappa)\log p(\mathbf{Y} |  (\mathbf{Z},\kappa))\mathrm{d}\mathbf{Z}\mathrm{d}\kappa\mathrm{d}\mathbf{Y}\\
&+\mathbf{H}(\mathbf{Y})
\end{aligned}
\end{equation}

$\mathbf{H}(\mathbf{Y})$ is is the entropy associated with $\mathbf{Y}$, which is irrelevant to the optimization objective and can therefore be ignored. 
However, since estimating $p(\mathbf{Y} |  (\mathbf{Z},\kappa))$ is challenging, it is common to use a variational approximation $q(\mathbf{Y} |  (\mathbf{Z},\kappa))$ as a substitute.

Due to the non-negativity of the KL scattering $\mathrm{KL}\left[p(y|z),q(y|z)\right]\geq0$, one gets:

\begin{equation}
    \int p(y|z)\log p(y|z)\mathrm{d}y\geq\int p(y|z)\log q(y|z)\mathrm{d}y
\end{equation}

Thus, Eq\eqref{eq:curvib_first} can be estimated:
\end{proof}

\subsection{Proof of Proposition~\ref{prop:upper_bound}}
We next provide the proof of Proposition~\ref{prop:upper_bound} in Section~\ref{sec:CurvGIB}.
\begin{proof}
According to the definition of mutual information, 
\begin{equation}
\begin{aligned}
\label{eq:curvib_first}
I(\mathbf{Z} | \kappa;\mathbf{Y})=&\iiint p((\mathbf{Z},\mathbf{Y}) | \kappa)\log p(\mathbf{Y} |  (\mathbf{Z},\kappa))\mathrm{d}\mathbf{Z}\mathrm{d}\kappa\mathrm{d}\mathbf{Y}\\&+\mathbf{H}(\mathbf{Y})
\\\approx& \iiint p((\mathbf{Z},\mathbf{Y}) | \kappa)\log p(\mathbf{Y} |  (\mathbf{Z},\kappa))\mathrm{d}\mathbf{Z}\mathrm{d}\kappa\mathrm{d}\mathbf{Y}
\\\ge& \iiint p((\mathbf{Z},\mathbf{Y}) | \kappa)\log q(\mathbf{Y} |  (\mathbf{Z},\kappa))\mathrm{d}\mathbf{Z}\mathrm{d}\kappa\mathrm{d}\mathbf{Y}
\end{aligned}
\end{equation}

For the second part of the  Eq\eqref{eq:curvib}:

\begin{equation}
\begin{aligned}
\label{eq:curvib_first}
I(\mathbf{Z} | \kappa;\mathbf{X})&=\iint p(\mathbf{Z} | \kappa;\mathbf{X})\log\frac{p(\mathbf{Z} | \kappa;\mathbf{X})}{p(\mathbf{Z} | \kappa)p(\mathbf{X})}\mathrm{d}(\mathbf{Z} | \kappa)\mathrm{d}\mathbf{X}
\\&=\iint p(\mathbf{Z} | \kappa;\mathbf{X})\log\frac{p( (\mathbf{Z} | \kappa) |  \mathbf{X})}{p(\mathbf{Z} | \kappa)}\mathrm{d}(\mathbf{Z} | \kappa)\mathrm{d}\mathbf{X}
\\&= \! \! \iiint \! p((\mathbf{Z}) | (\kappa,\!\mathbf{X}\!)) \! \log\! \frac{p(\mathbf{Z} |  (\mathbf{X},\kappa))}{p(\mathbf{Z} | \kappa)}\mathrm{d}\mathbf{Z}\mathrm{d}\kappa\mathrm{d}\mathbf{X}
\end{aligned}
\end{equation}

Similarly, since estimating $p(\mathbf{Z} | \kappa)$ is difficult, we introduce an  variational approximation $r(\mathbf{Z} | \kappa)$ as a substitute.

\begin{equation}
\begin{aligned}
\label{eq:curvib_first}
I(\mathbf{Z} | \kappa;\!\mathbf{X}\!)=& \!\! \iiint \!\! p((\mathbf{Z}) | (\kappa,\!\mathbf{X}\!))\log\frac{p(\mathbf{Z} |  (\mathbf{X},\kappa))}{p(\mathbf{Y})}\mathrm{d}\mathbf{Z}\mathrm{d}\kappa\mathrm{d}\mathbf{X}
\\ \!\le&\! \int\!\! p(\mathbf{X})p(\mathbf{Y}|(\mathbf{X},\kappa))\mathrm{log}\frac{p(\mathbf{Y}|(\mathbf{X},\kappa))}{r(\mathbf{Z}|\kappa)}d\kappa d\mathbf{X}d\mathbf{Z},
\end{aligned}
\end{equation}

\end{proof}

\section{Experiment Details}

We evaluate \modelname
% \footnote{Code is available at \url{https://github.com/RingBDStack/VIB-GSL}. }
~on two tasks: node classification and graph denoising, to verify whether \modelname~can improve the effectiveness and robustness of graph representation learning. 
Then we analyze the impact of information compression quantitatively and qualitatively. 
\subsection{Experimental Setups}
\subsubsection{Datasets. }
% \footnote{Available at https://chrsmrrs.github.io/datasets/docs/datasets/}
We empirically perform experiments on eight widely-used real datasets including Karate Club\cite{Avrachenkov1977AnIF}, Cora, Citeseer, Pubmed~\cite{Sen2008CollectiveCI}, Coauthor-CS,Coauthor-Physics~\cite{shchur2019pitfallsgraphneuralnetwork},Amazon-Photos,Amazon-Computers ~\cite{Rossi2015TheND}. 
We choose these datasets for evaluation because much noisy information may exist in real world interactions. 
% Please refer to Appendix~\ref{app:dataset} for dataset statistics. 
% IMDB-B and IMDB-M are . 
% REDDIT-B is. 
% COLLAB is .
\subsubsection{Baselines. }
We compare the proposed \modelname~with a number of node-level  baselines which can be devided into three categories, including ricci method (BORF~\cite{Nguyen2022RevisitingOA}, CurvGN~\cite{Ye2020CurvatureGN}) , structure learning method( SUBLIME~\cite{Liu2022TowardsUD}, MVPOOL\_SL ~\cite{Zhang2021HierarchicalMG}) and information bottleneck method (RGIB ~\cite{Wang2022TowardER}), to demonstrate the effectiveness and robustness of \modelname. 
We do not include VIB-GSL in our baselines since it focuses on graph-level representation learning. 

\subsection{Time Complexity}
\label{app:time}
\vspace{+1em}
% \section{Model Complexity}
The computational complexity of our model can be determined in two parts. 
The first part is to compress the node representation by information bottleneck, and the second part is to update the curvature metric to refine the structures.

\noindent For \modelname, the cost of the first part is $\mathcal{O}(nd+n^2d)$ for a graph with $n$ nodes in $\mathbf{R}^{d}$ where $d$ is the node feature dimension.And the cost of secoond part is $\mathcal{O}(n{d^2})$ where $n$ is the num of nodes and 
$d$ is the node feature dimension.
If we assume that  $d\ll n$, the overall time complexity is $\mathcal{O}(n^2)$. 
\end{document}